\documentclass[conference]{IEEEtran}
\IEEEoverridecommandlockouts
\usepackage{cite}
\usepackage{amsmath,amssymb,amsfonts}
\usepackage{graphicx}
\usepackage{textcomp}
\usepackage{xcolor}
\usepackage{multirow}

\usepackage{hyperref}
\usepackage{algorithm}
\usepackage{algpseudocode}

\algtext*{EndFor}
\algtext*{EndWhile}
\algtext*{EndIf}

\def\BibTeX{{\rm B\kern-.05em{\sc i\kern-.025em b}\kern-.08em
    T\kern-.1667em\lower.7ex\hbox{E}\kern-.125emX}}
\begin{document}

\title{M3R: Localized Rainfall Nowcasting with Meteorology-Informed MultiModal Attention}

\author{
\IEEEauthorblockN{Sanjeev Panta$^{\star}$, Rhett M Morvant$^{\star}$, Xu Yuan$^{\dagger}$, Li Chen$^{\star}$, Nian-Feng Tzeng$^{\star}$}
\IEEEauthorblockA{$^{\star}$University of Louisiana at Lafayette, Lafayette, LA, USA\\
$^{\dagger}$University of Delaware, Newark, DE, USA\thanks{The research is supported in part by the NSF under grants OIA-2327452, OIA-2019511, and 2425812, in part by the Louisiana BoR under LEQSF(2024-27)-RD-B-03. Corresponding author: Dr. Li Chen (li.chen@louisiana.edu)}}
}

\maketitle

\begin{abstract}
Accurate and timely rainfall nowcasting is crucial for disaster mitigation and water resource management. Despite recent advances in deep learning, precipitation prediction remains challenging due to limitations in effectively leveraging diverse multimedia data sources. We introduce M3R, a Meteorology-informed MultiModal attention-based architecture for direct Rainfall prediction that synergistically combines visual NEXRAD radar imagery with numerical Personal Weather Station (PWS) measurements, using a comprehensive pipeline for temporal alignment of heterogeneous meteorological data. With specialized multimodal attention mechanisms, M3R novelly leverages weather station time series as queries to selectively attend to spatial radar features, enabling focused extraction of precipitation signatures. Experimental results for three spatial areas of 100 km $\times$ 100 km centered at NEXRAD radar stations demonstrate that M3R outperforms existing approaches, achieving substantial improvements in accuracy, efficiency, and precipitation detection capabilities. Our work establishes new benchmarks for multimedia-based precipitation nowcasting and provides practical tools for operational weather prediction systems. The source code is available at \href{https://github.com/Sanjeev97/M3Rain}{https://github.com/Sanjeev97/M3Rain}
\end{abstract}

\begin{IEEEkeywords}
multimedia, multimodality, transformer, time series
\end{IEEEkeywords}
\section{Introduction}

\label{sec:intro}
\begin{figure*}[!t]
\centering
\includegraphics[height=4.8cm, width=\textwidth]{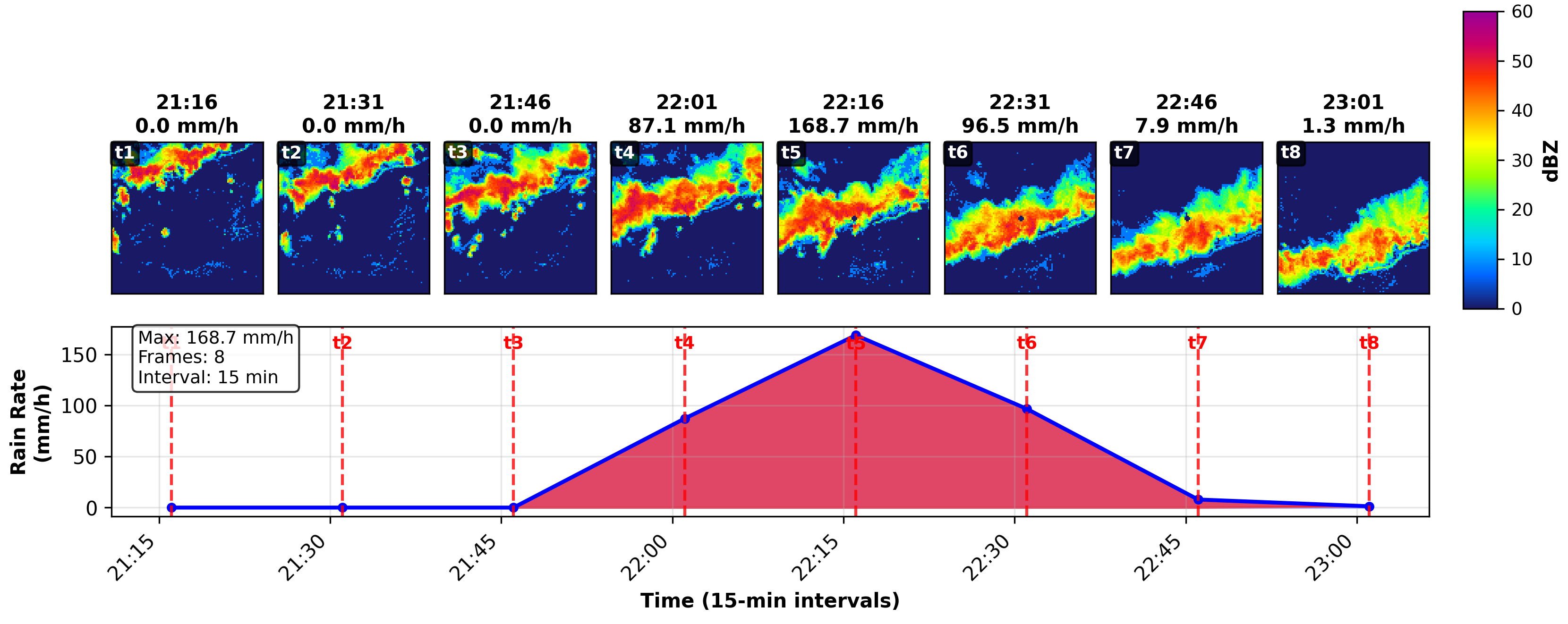}
\caption{Rainfall Event Sample (Sequence 5655) of the LA dataset.}
\label{fig:rain_event}
\end{figure*}

Rainfall nowcasting, the short-term prediction of precipitation within a lead time of up to six hours, is vital for mitigating weather-related hazards and supporting timely decision-making in flood control, transportation, and emergency services. Traditional approaches rely on single-media sources, primarily NEXRAD radar imagery, which suffers from inherent limitations including ground clutter, beam blockage, and systematic uncertainties from reflectivity-to-precipitation conversion \cite{gao2023prediff, yu2024diffcast}.
These single-media approaches fail to capture the full complexity of precipitation phenomena and often struggle with accuracy in localized prediction scenarios.

Integrating additional data modalities or multimedia presents significant potential to address these limitations. Ground-based weather station measurements provide direct and unambiguous precipitation observations that can serve as crucial ground-truth constraints for radar interpretation. However, while weather stations offer precise point measurements, they lack the spatial coverage necessary for understanding regional precipitation patterns and dynamics. This complementary relationship, where radar provides broad spatial context and weather stations offer accurate ground measurements, presents a compelling opportunity for multimodal learning approaches that can leverage the strengths of multimedia data sources while mitigating their respective limitations.

Precipitation nowcasting has evolved from traditional methods to sophisticated deep learning approaches\cite{lin2023comprehensive}. Initial advances combined recurrent and convolutional architectures: ConvLSTM \cite{shi2015convolutional} embedded spatial convolutions within LSTM cells for radar extrapolation, while TrajGRU \cite{shi2017deep} introduced trajectory-based gating with learnable motion representations. Subsequent work adapted U-Net encoder-decoder architectures for continental-scale applications, including MRMS \cite{agrawal2019machine} and SEVIR \cite{veillette2020sevir} for large-scale radar datasets.

Recent time series architectures \cite{zeng2023transformers, nie2022time, panta2026revisiting, liu2023itransformer} leveraging linear model, patches and inverted dimensions, limit to numerical weather data \cite{zhang2025regional, zhang2021precise}. Radar-based transformer approaches include Earthformer \cite{gao2022earthformer} with space-time transformers using cuboid attention. Generative models emerged including PreDiff \cite{gao2023prediff} with latent diffusion, DiffCast \cite{yu2024diffcast} using residual diffusion to decompose precipitation into global deterministic motion and local stochastic variations, and AlphaPre\cite{lin2025alphapre} with amplitude-phase disentanglement model.


Limited works have explored multimedia dataset for precipitation nowcasting. Some multimodal approaches are explored in other domains like agriculture for crop yield predictions\cite{lin2023mmst, lin2024open}. MM-RNN \cite{ma2023mm} introduced multimodal RNNs with  modal fusion mechanisms, while FsrGAN \cite{niu2024fsrgan} addressed satellite-radar fusion through encoder-fusion-decoder networks with spatial-channel attention. METEO-DLNet \cite{hu2024meteo} incorporated meteorological variables alongside radar data using multi-scale processing before converting reflectivity to rainfall. They treat multimedia datasets symmetrically, failing to recognize the fundamental meteorological principle that local point measurements are influenced by broader spatial weather patterns. Thus, we are motivated to develop a multimedia architecture that naturally encodes the asymmetric relationship between spatial radar patterns and temporal weather station measurements while providing direct precipitation predictions without the need for computationally intensive reflectivity-precipitation conversion \cite{shi2015convolutional}.

More specifically, our core insight is that weather station time series should query spatial radar features to identify relevant atmospheric patterns, mimicking how meteorologists interpret localized measurements within broader spatial contexts. As such, our framework employs an asymmetric attention mechanism to enable selective focus on radar regions most relevant to specific ground measurements, while simultaneously providing direct quantitative precipitation outputs that eliminate conversion uncertainties. Intuitively, this approach benefits from combining the comprehensive spatial awareness of radar with the precision of ground measurements in a meteorologically informed manner, resulting in more accurate and efficient precipitation nowcasting compared to existing single-modal or symmetrically-fused multimodal methods. 

In addition, we address the critical need for Multimedia or MultiModal Rainfall datasets with sufficient spatial resolution to support localized forecasting in small-scale regions, with our effort on preparing new datasets (Lake Charles, Jackson, Montgomery) based on NOAA NEXRAD LEVEL 2 Base dataset \cite{huber2009review} and Personal Weather Station dataset\footnote{https://www.wunderground.com/}, comprising image-based and numerical modalities.

In summary, current approaches suffer from fundamental limitations: (1) \textbf{indirect precipitation prediction} requiring Z-R conversion, introducing uncertainty and overhead; (2) \textbf{limited multimedia integration} failing to effectively leverage complementary information; (3) \textbf{multimedia dataset limitations} targeting large-scale prediction with limited applicability to localized nowcasting; (4) \textbf{methodological gaps} lacking comprehensive evaluation against fundamental baselines and detailed multimodal fusion analysis.


We propose M3R, a multimedia transformer architecture addressing these limitations through the following contributions:

    \textbf{Meteorology-Informed Multimedia Architecture}: Novel Multimedia Transformer with a meteorology-informed multimodal attention enabling time series queries to selectively attend to spatial radar patterns, improving computational efficiency.
    
    \textbf{Direct Prediction Framework}: Direct quantitative precipitation outputs eliminating Z-R conversion uncertainties and computational overhead.
    
    \textbf{Dataset Processing Pipeline}: Systematic framework for processing and temporally aligning heterogeneous multimedia meteorological datasets with novel rainfall event selection algorithm for high-quality training data curation. Our pipeline handles the complex challenges of synchronizing irregular radar observations with weather station measurements.
    
    \textbf{Localized-Scale Forecasting}: Effective prediction for small cities (e.g., Lake Charles, Jackson, Montgomery) at 15-minute granularity up to 1 hour ahead, addressing localized weather prediction needs.
    

 Experimental results demonstrate that our model achieves 12-27\% and 20-34\% improvements in prediction accuracy across three different locations, over the time series and the radar nowcasting baselines.    
\section{Multimedia Meteorological Dataset \& Processing Pipeline}
\label{section:dataset}

We develop a comprehensive processing pipeline for temporally aligning heterogeneous meteorological datasets to enable multimodal precipitation analysis. The datasets created for three areas in Lake Charles (LA), Montgomery (AL) and Jackson (MS), each covering a 100 km × 100 km spatial area, span January 2022 to September 2024 (33 months). Those three spatial domains are centered at the NEXRAD radar stations (KLCH, KMXX, KDGX) respectively. We integrate NOAA NEXRAD Level-2 radar data (8-20 minute intervals) with Personal Weather Station data recording 20 meteorological variables (5-7 minute intervals). The pipeline produces 96,359 instances of aligned reflectivity-meteorological data at 15-minute granularity, yielding 7044, 6045 and 9122 high-quality weather event sequences (e.g., Fig. \ref{fig:rain_event}) with 8 temporal frames each (56352, 48360, 72976 total). Our key innovation lies in the sophisticated temporal alignment and event selection methodology.

\textbf{NEXRAD Processing:} The radar processing pipeline transforms 3D radar volumes into regular 2D composite reflectivity fields. Raw NEXRAD Level-2 files are converted to NetCDF format (using LROSE\footnote{http://lrose.net/}) with coordinate transformation to Cartesian grids using Lambert Conformal Conic projection at 1 km resolution. Column-maximum composite reflectivity uses the four lowest elevation angles to minimize beam blockage:
\begin{align}
\mathbf{Z}_c(x, y) = \max_{k=1}^{4} \mathbf{Z}_k(x, y)
\end{align}
Temporal regularization employs piecewise linear interpolation to create uniform 15-minute intervals:
\begin{align}
\mathbf{Z}^{\prime}(x, y, t_j) = \mathbf{Z}(x, y, t_i) + \frac{\mathbf{Z}(x, y, t_{i+1}) - \mathbf{Z}(x, y, t_i)}{t_{i+1} - t_i} \cdot (t_j - t_i)
\end{align}

\textbf{PWS Processing: }
PWS data undergo quality control with unit standardization, temporal alignment with radar data to 15-minute grids, and gap filling. We retain 20 meteorological variables encompassing thermal, moisture, dynamic, and precipitation measurements. Wind data employ vector decomposition to handle directional discontinuities
followed by interpolation and reconstitution. Precipitation receives contextual interpolation through rolling window analysis within $\tau = 2.5$ hours.

\textbf{MultiModal Alignment and Event Selection:} We identify meteorologically significant precipitation episodes through a systematic algorithm shown in Algorithm \ref{alg:event_selection}. 
\begin{align} \label{eqn:quantization}
\mathbf{Q(Z_r)} = \begin{cases}
0 & \text{if } \mathbf{Z} < 8 \\
8 & \text{if } 8 \leq \mathbf{Z} < 16 \\
16 & \text{if } 16 \leq \mathbf{Z} < 20 \\
\lfloor \mathbf{Z} \rfloor & \text{if } 20 \leq \mathbf{Z} < 70 \\
70 & \text{if } \mathbf{Z} \geq 70 \\
255 & \text{if } \mathbf{Z} \text{ is missing}
\end{cases}
\end{align}
This scheme $\mathbf{Q(Z_r)}$ applies different resolutions based on meteorological significance: coarse binning for light precipitation (8-20 dBZ) and fine resolution for convective echoes (20-70 dBZ). \textbf{Please see the supplementary material for details.}


\begin{algorithm}[t]
\caption{Meteorological Event Selection \& Alignment}
\label{alg:event_selection}
\small
\begin{algorithmic}[1]
\Require Raw radar reflectivity $\mathbf{Z}_r^{(t)}$ for $t = 1, \ldots, T$, and threshold $\bar{Z}_{\text{thres}} = 3.0$ dBZ
\Ensure Event sequence set $\mathcal{E}$ with aligned PWS data
\For{each time $t$}
    \State Apply quantization (Eq.~\ref{eqn:quantization}): $\mathbf{Z}^{(t)} \leftarrow Q(\mathbf{Z}_r^{(t)})$
\EndFor
\State $\mathcal{E} \leftarrow \emptyset$, $i \leftarrow 4$, $\Delta t = 15$ min
\While{$i + 4 < T$}
    \State $\bar{Z}_i \leftarrow \frac{1}{N_x \cdot N_y} \sum_{x,y} Z_{x,y}^{(t_i)}$ \Comment{Spatial mean at time $t_i$}
    \If{$\bar{Z}_i > \bar{Z}_{\text{thres}}$}
        \State $W_i \leftarrow \{\mathbf{Z}^{(t_{i+j})} \mid j \in [-4, 3]\}$ \Comment{8-frame window}
        \State $\mathcal{E} \leftarrow \mathcal{E} \cup \{(W_i, \{t_{i+j}\}_{j=-4}^{3})\}$
        \State $i \leftarrow i + 4$ \Comment{Non-overlapping stride}
        \State \textbf{continue}
    \EndIf
    \State $i \leftarrow i + 4$
\EndWhile
\For{each $(W_i, \{t_j\})$ in $\mathcal{E}$}
    \For{each timestamp $t_j$}
        \State $t_{\text{PWS}}^* \leftarrow \arg\min_{t_{\text{PWS}}} |t_j - t_{\text{PWS}}|$ \Comment{PWS timestamp}
        \State Store aligned pair $(\mathbf{Z}^{(t_j)}, \mathbf{P}^{(t_{\text{PWS}}^*)})$
    \EndFor
\EndFor
\State \Return $\mathcal{E}$
\end{algorithmic}
\end{algorithm}

\section{Model Architecture}
This section presents the Multimedia Model (M3R) architecture, a novel deep learning framework specifically designed for multimedia meteorological prediction tasks. The architecture integrates spatiotemporal radar imagery with concurrent numerical meteorological time series data through an innovative multimodal attention mechanism.

\textbf{Problem Formulation}: Consider a sequence of radar reflectivity images $\mathbf{X} \in \mathbb{R}^{T \times H \times W  \times C}$ and the corresponding meteorological time series $\mathbf{Z} \in \mathbb{R}^{T \times D}$, where, $T$ denotes temporal sequence length,  $H \times W$ are spatial dimensions, $C$ is the number of radar channels (set to 1 as we are using composite of first four layers of reflectivities), and $D$ is the feature dimensionality of meteorological variables. 
The objective is to learn a mapping function $f : (\mathbf{X}, \mathbf{Z}) \rightarrow \mathbf{Y}$, 
where $\mathbf{Y} \in \mathbb{R}^{T \times 1}$ represents predicted precipitation rates.

\begin{figure}[t]
\centering
\includegraphics[height=9cm, width=0.8\columnwidth]{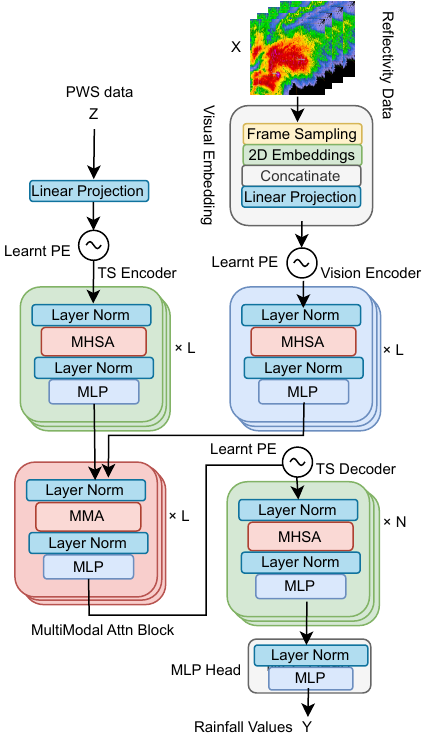}
\caption{Illustration of M3R Model with Vision Encoder, TS Encoder, MultiModal Attention Block, TS Decoder and MLP Head. 
} \label{fig:architecture}
\end{figure}

\textbf{Architecture Overview:} The proposed M3R architecture (Fig.~\ref{fig:architecture}) comprises the following main components: Embeddings, Encoders, MultiModal Block, and TS Decoder.
\begin{figure}[t]
\centering
\includegraphics[width=0.7\columnwidth]{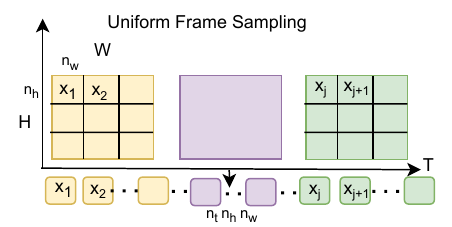}
\caption{Illustration of Uniform Frame Sampling for Visual Embedding.
} \label{fig:architecture-sampling}
\end{figure}
\subsection{Radar Patch Embedding}
Following the Vision Transformer paradigm for Visual Embedding \cite{alexey2020image}, radar images are decomposed into patches. We incorporate uniform frame sampling technique \cite{arnab2021vivit} to treat images in each timestamp as frames of a video clip, embed each 2D frame independently using the same method as ViT \cite{alexey2020image} and concatenate all these tokens together to process with the vision encoder. 

Given input radar sequence 
$\mathbf{X} \in \mathbb{R}^{T \times H  \times W \times C}$, we 
independently tokenize each temporal frame into non-overlapping spatial 
patches of size $P \times P$. If $n_h . n_w$ image patches are extracted from each frame, as in \cite{alexey2020image}, then a total of $n_t . n_h . n_w$ tokens will be forwarded as in Fig. \ref{fig:architecture-sampling}. Intuitively, this process is seen as simply constructing a large 2D image to be tokenised following ViT. Each frame is independently partitioned 
into $N_{\text{patches}} = \frac{H \cdot W}{P^2}$:
\label{eq:patch_extraction}
\begin{align}  
\mathbf{X}_{\text{patches}} = \text{Rearrange}(\mathbf{X}, 
(T, \tfrac{H}{P} \cdot \tfrac{W}{P}, C \cdot P^2))
\end{align}
where $P$ represents the patch size. Each patch is then linearly projected to the model dimension $d_\text{model}$ such that radar context embedding $\mathbf{E}_\text{ctx} \in \mathbb{R}^{T \times N_\text{patches} \times d_\text{model}}$:
\begin{align}
\mathbf{E}_{\text{ctx}} = \mathbf{X}_{\text{patches}} \mathbf{W}_{\text{patch}} + \mathbf{b}_{\text{patch}}
\end{align}
where $\mathbf{W}_{\text{patch}} \in \mathbb{R}^{C \cdot P^2 \times d_{\text{model}}}$ and $\mathbf{b}_{\text{patch}} \in \mathbb{R}^{d_{\text{model}}}$ are learnable parameters. We add learnt positional encoding \cite{alexey2020image, arnab2021vivit} $\mathbf{PE}_\text{ctx} \in \mathbb{R}^{N_{\text{patches}} \times d_{\text{model}}}$ to preserve spatial relationships:
\begin{align}
\tilde{\mathbf{E}}_{\text{ctx}} = \mathbf{E}_{\text{ctx}} + \mathbf{PE}_{\text{ctx}}
\end{align}

\subsection{Meteorological Time Series Embedding}
The meteorological time series $\mathbf{Z} \in \mathbb{R}^{T \times D}$ undergoes linear projection to align with the model dimension such that $\mathbf{E_{ts}} \in \mathbb{R}^{T \times d_\text{model}}$:
\begin{align}
\mathbf{E}_{\text{ts}} = \mathbf{Z} \mathbf{W}_{\text{ts}} + \mathbf{b}_{\text{ts}}
\end{align}
where $\mathbf{W}_{\text{ts}} \in \mathbb{R}^{D \times d_{\text{model}}}$ and $\mathbf{b}_{\text{ts}} \in \mathbb{R}^{d_{\text{model}}}$ are learnable transformation matrices. Learnt positional encoding $\mathbf{PE}_{\text{ts}} \in \mathbb{R}^{T \times d_{\text{model}}}$ is incorporated, following the patch embedding instead of the absolute positional encoding from the vanilla transformer \cite{vaswani2017attention}:
\begin{align}
\tilde{\mathbf{E}}_{\text{ts}} = \mathbf{E}_{\text{ts}} + \mathbf{PE}_{\text{ts}}
\end{align}
\subsection{Encoders}
Both radar and time series embeddings are independently processed through multi-head self-attention layers. For the Radar Context Encoder or Vision Encoder:
\begin{align}
\mathbf{H}_{\text{ctx}}^{\prime(l)} &= \text{MHSA}(\text{LN}(\mathbf{H}_{\text{ctx}}^{(l-1)})) + \mathbf{H}_{\text{ctx}}^{(l-1)}, \mathbf{H}_{\text{ctx}}^{(0)} = \tilde{\mathbf{E}}_{\text{ctx}} \\
\mathbf{H}_{\text{ctx}}^{(l)} &= \text{MLP}(\text{LN}(\mathbf{H}_{\text{ctx}}^{\prime(l)})) + \mathbf{H}_{\text{ctx}}^{\prime(l)} 
\end{align} 
where $\text{MHSA}(\cdot)$ denotes multi-head self-attention, $\text{LN}(\cdot)$ represents layer normalization, $\text{MLP}(\cdot)$ is a multi-layer perceptron, and $l$ indexes the layer.
And the multi-head self-attention mechanism \cite{vaswani2017attention} is computed as:
\begin{align}
\text{MHSA}(\mathbf{H}) = \text{Concat}(\text{head}_1, \ldots, \text{head}_h) \mathbf{W}^O
\end{align}
where each attention head is defined as:
\begin{align}
\text{head}_i = \text{Attention}(\mathbf{H}\mathbf{W}_i^Q, \mathbf{H}\mathbf{W}_i^K, \mathbf{H}\mathbf{W}_i^V) \\
\text{Attention}(\mathbf{Q}, \mathbf{K}, \mathbf{V}) = \text{softmax}\left(\frac{\mathbf{Q}\mathbf{K}^T}{\sqrt{d_k}}\right)\mathbf{V}
\end{align}
with projection matrices $\mathbf{W}_i^Q, \mathbf{W}_i^K, \mathbf{W}_i^V \in \mathbb{R}^{d_\text{model} \times d_k}$  where $d_k = d_{\text{model}}/h$ is the head dimension and $h$ is the number of attention heads, and output projection 
$\mathbf{W}^O \in \mathbb{R}^{h \cdot d_k \times d_\text{model}}$,
Similarly, the Time Series (TS) Encoder processes $\tilde{\mathbf{E}}_{\text{ts}}$ through analogous self-attention layers to produce $\mathbf{H}_{\text{ts}}^{(L)}$.
\subsection{MultiModal Block}
The core innovation of M3R architecture lies in its meteorology-informed multimodal attention mechanism. This design reflects the fundamental principle that local point measurements (from weather station) are influenced by broader spatial weather patterns (radar imagery). We implement this through an attention mechanism where time series representations selectively query radar features, mimicking how meteorologists interpret ground measurements within the context of spatial precipitation patterns.
\begin{align}
\mathbf{H}_{\text{mm}}^{\prime(l)} &= \text{MultiModalAttn}(\mathbf{H}_{\text{ctx}}^{(L)}, \mathbf{H}_{\text{mm}}^{(l-1)}) + \mathbf{H}_{\text{mm}}^{(l-1)}, \mathbf{H}_{\text{mm}}^{(0)} = \mathbf{H}_{\text{ts}}^{(L)} \\
\mathbf{H}_{\text{mm}}^{(l)} &= \text{MLP}(\text{LN}(\mathbf{H}_{\text{mm}}^{\prime(l)})) + \mathbf{H}_{\text{mm}}^{\prime(l)}
\end{align}
The multimodal attention mechanism is defined as:
\begin{equation}
\text{MultiModalAttn}(\mathbf{H}_{\text{src}}, \mathbf{H}_{\text{tgt}}) = \text{Concat}(\text{head}_1, \ldots, \text{head}_h)\mathbf{W}^O
\end{equation}
\begin{equation}
\text{head}_i = \text{Attention}(\mathbf{H}_{\text{tgt}}\mathbf{W}_i^Q, \mathbf{H}_{\text{src}}\mathbf{W}_i^K, \mathbf{H}_{\text{src}}\mathbf{W}_i^V)
\end{equation}
where each multimodal attention head computes head$_i$.
This formulation allows the time series queries $\mathbf{H}_{\text{tgt}}\mathbf{W}_i^Q$ to attend to relevant spatial patterns in radar-derived keys $\mathbf{H}_{\text{src}}\mathbf{W}_i^K$ and values $\mathbf{H}_{\text{src}}\mathbf{W}_i^V$, reflecting the physical relationship between spatial precipitation patterns and their manifestation at ground measurement stations.
\subsection{Temporal Prediction Decoder}
The fused representation $\mathbf{H}_{\text{mm}}$ is processed through a temporal transformer decoder or TS Decoder to generate precipitation predictions. 
\begin{align}
\mathbf{H}_{\text{dec}}^{\prime(l)} &= \text{MHSA}(\text{LN}(\mathbf{H}_{\text{dec}}^{(l-1)})) + \mathbf{H}_{\text{dec}}^{(l-1)},    \mathbf{H}_{\text{dec}}^{(0)} = \mathbf{H}_{\text{mm}}^{(L)} \\
\mathbf{H}_{\text{dec}}^{(l)} &= \text{MLP}(\text{LN}(\mathbf{H}_{\text{dec}}^{\prime(l)})) + \mathbf{H}_{\text{dec}}^{\prime(l)}
\end{align}
where $\mathbf{H}_{\text{dec}}^{\prime(l)} \in \mathbb{R}^{T \times d_\text{dec}}$ and $\text{MHSA}(.)$ denotes multi-head self-attention. The final prediction is obtained through a linear projection layer:
\begin{align}
\mathbf{Y} = \mathbf{H}_{\text{dec}}^{(L_{\text{dec}})}\mathbf{W}_{\text{out}} + \mathbf{b}_{\text{out}}
\end{align}
where $\mathbf{W}_{\text{out}} \in \mathbb{R}^{d_{\text{dec}} \times 1}$ and $\mathbf{b}_{\text{out}} \in \mathbb{R}^{1}$ are the output projection parameters.

\section{Experiment}

\begin{table*}[!h]
\caption{Comparison showing our model is \textbf{best} or \underline{second best} in both time series forecasting and binary rain classification over three spatial domains of interest (denoted by LA, AL and MS).}
\centering
\resizebox{\textwidth}{!}{
\begin{tabular}{|c|ccc|ccc|ccc|ccc|ccc|ccc|ccc|}
\hline
\textbf{Methods} & \multicolumn{3}{c|}{\textbf{RMSE} ($\downarrow$)} & \multicolumn{3}{c|}{\textbf{MAE} ($\downarrow$)} & \multicolumn{3}{c|}{\textbf{R$^2$} ($\uparrow$)} & \multicolumn{3}{c|}{\textbf{CC} ($\uparrow$)} & \multicolumn{3}{c|}{\textbf{CSI 0.1} ($\uparrow$)} & \multicolumn{3}{c|}{\textbf{CSI 5} ($\uparrow$)} & \multicolumn{3}{c|}{\textbf{CSI 10} ($\uparrow$)} \\
\cline{2-22}
& \textbf{LA} & \textbf{AL} & \textbf{MS} & \textbf{LA} & \textbf{AL} & \textbf{MS} & \textbf{LA} & \textbf{AL} & \textbf{MS} & \textbf{LA} & \textbf{AL} & \textbf{MS} & \textbf{LA} & \textbf{AL} & \textbf{MS} & \textbf{LA} & \textbf{AL} & \textbf{MS} & \textbf{LA} & \textbf{AL} & \textbf{MS} \\
\hline
\hline
Transformer \cite{vaswani2017attention} & 3.79 & 4.06 & 3.68 & 0.74 & 1.03 & 0.64 & 0.01 & 0.004 & 0.09 & 0.12 & 0.06 & 0.30 & 0.060 & 0.071 & 0.070 & 0.174 & 0.029 & 0.221 & 0.022 & 0.024 & 0.172 \\
\hline
DLinear \cite{zeng2023transformers} & 3.27 & 3.10 & 3.49 & 0.81 & 0.83 & 0.77 & 0.06 & 0.08 & 0.12 & 0.25 & 0.28 & 0.35 & 0.047 & 0.051 & 0.042 & 0.211 & 0.125 & 0.234 & 0.114 & 0.037 & 0.154 \\
\hline
PatchTST \cite{nie2022time} & 3.63 & 3.14 & 3.56 & 0.45 & 0.36 & 0.41 & 0.07 & 0.06 & 0.10 & 0.26 & 0.24 & 0.32 & 0.366 & 0.359 & 0.372 & 0.226 & 0.145 & 0.340 & 0.133 & 0.071 & 0.250 \\
\hline
iTransformer \cite{liu2023itransformer} & 3.43 & 3.41 & 3.54 & 0.49 & 0.41 & 0.41 & 0.08 & 0.02 & 0.13 & 0.28 & 0.15 & 0.36 & 0.365 & 0.349 & 0.355 & 0.233 & 0.074 & 0.339 & 0.125 & 0.066 & 0.213 \\
\hline
\hline
Diffcast-SimVP \cite{yu2024diffcast} & 4.70 & 3.13 & 4.14 & 0.97 & 0.60 & 0.65 & 0.05 & 0.09 & 0.08 & 0.21 & 0.31 & 0.29 & 0.074 & 0.092 & 0.073 & 0.218 & 0.114 & 0.377 & 0.105 & 0.100 & \textbf{0.284} \\
\hline
Diffcast-PhyDnet \cite{yu2024diffcast} & 3.84 & 3.40 & \textbf{3.27} & 0.54 & 0.67 & 0.64 & 0.02 & 0.06 & \textbf{0.25} & 0.13 & 0.25 & \textbf{0.50} & 0.136 & 0.091 & 0.072 & \textbf{0.256} & 0.148 & 0.391 & 0.060 & \textbf{0.170} & 0.268 \\
\hline
AlphaPre \cite{lin2025alphapre} & 3.10 & \textbf{2.94} & 3.35 & 0.43 & 0.52 & 0.55 & 0.12 & \textbf{0.16} & 0.19 & 0.35 & \textbf{0.40} & 0.44 & 0.150 & 0.092 & 0.075 & 0.163 & \textbf{0.210} & 0.386 & 0.000 & 0.140 & 0.176 \\
\hline
\hline
M3R (Ours) & \textbf{2.87} & \underline{3.03} & \underline{3.28} & \textbf{0.33} & \textbf{0.36} & \textbf{0.36} & \textbf{0.29} & \underline{0.11} & \underline{0.23} & \textbf{0.54} & \underline{0.34} & \underline{0.48} & \textbf{0.410} & \textbf{0.300} & \textbf{0.414} & \underline{0.240} & 0.109 & \textbf{0.407} & \textbf{0.236} & 0.130 & 0.264 \\
\hline
\end{tabular}
}
\label{tab:main_results}
\end{table*}

\begin{figure*}[htbp]
\includegraphics[height=3.2cm, width=\textwidth]{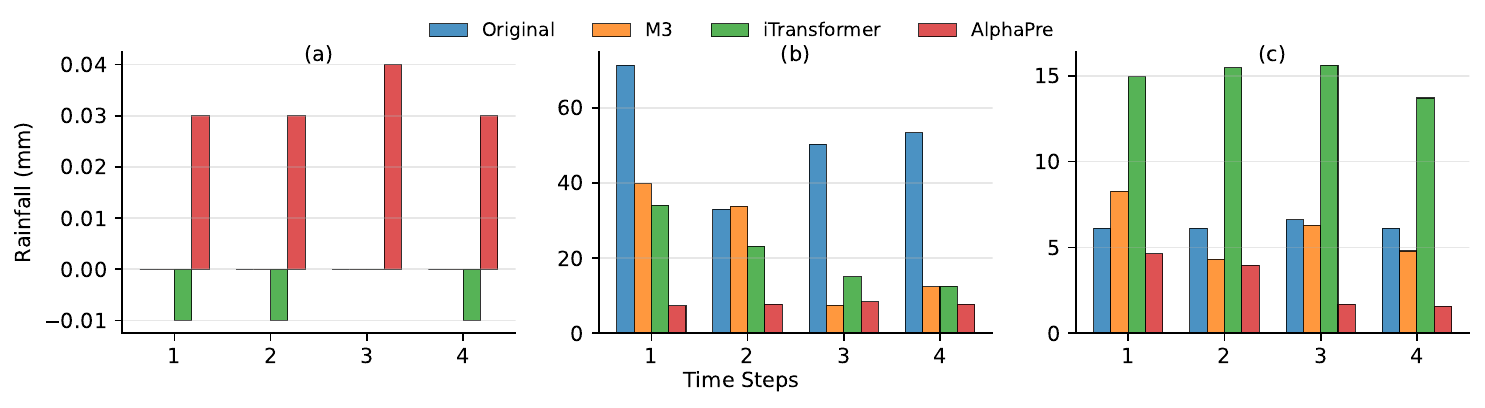}
\vspace*{-5mm}
\caption{Comparison: (a) Zero (b) High-intensity and (c) Medium-intensity precipitation showing M3R's consistent performance across all scenarios for LA.}
\label{fig:comparison}
\end{figure*}
\subsection{Experimental Setup}
We utilize the datasets (LA, AL, MS) produced by pipeline in Section \ref{section:dataset}. We use 85:15 chronological train-test split with prediction horizon and input window of 4 frames (1 hour). 

We train M3R for 200 epochs on single NVIDIA GPU with batch size 64, learning rate 1e-3, cosine annealing scheduler, 20 warmup epochs, AdamW optimizer with weight decay 0.05, and Mean Squared Error (MSE) loss function. 
We apply precipitation significance threshold 3.0 dBZ for sequence selection. Implementation uses PyTorch with mixed precision training. For Diffcast and AlphaPre baselines, we converted PWS reflectivity values to rainfall using the Z-R relationship. (\textbf{Please see Supplementary Section II for model details}) Each experiment ran three times with averaged results. 
Performance evaluation uses root mean squared error (RMSE), mean absolute error (MAE), R-squared (R$^2$), Correlation Coefficient (CC), and Critical Success Index (CSI) with thresholds 0.1, 5 and 10 mm/hr on the test set.

\subsection{Results and Analysis}
M3R demonstrates exceptional performance across evaluation metrics, achieving substantial improvements over both time series forecasting and precipitation nowcasting approaches. As shown in Table \ref{tab:main_results}, M3R achieves the \textbf{best MAE across all stations}, representing 12-27\% MAE improvement over PatchTST for three spatial areas of interest . Most notably, M3R shows superior pattern recognition with highest and second highest correlation coefficients and R² values across all stations, \textbf{exhibiting 1.8-3.6× more explanatory power} than the best time series baselines. At Lake Charles (LA), for example: M3R achieves exceptional R² of 0.29 and MAE of 0.33 along with almost all other metrics, demonstrating that meteorological spatial radar context information added to numerical rainfall parameters significantly improves rainfall forecasting accuracy. Results for the other two areas Jackson (MS) and Montgomery (AL), exhibit similar improvements in forecasting accuracy. 
\textbf{Please see Supplementary Section III for detailed discussion.}

Comparison with state-of-the-art radar-based methods reveals M3R's substantial advantages. M3R achieves \textbf{consistently superior MAE} compared to AlphaPre with 20-34\% improvement across all locations, along with 1.2-2.4× higher R² and noticably higher CSI 0.1 for light precipitation detection across stations, with AlphaPre failing completely at heavy precipitation detection at LA (CSI 10: 0.000). Beyond accuracy, M3R demonstrates exceptional computational efficiency (Table \ref{tab:efficiency_analysis}) with \textbf{13× faster training} and \textbf{7× faster inference} than AlphaPre while maintaining lower FLOPs than radar-based methods. Unlike traditional approaches requiring Z-R conversion, M3R \textbf{directly outputs quantitative precipitation values}, eliminating conversion uncertainties while delivering consistent performance across geographically diverse stations.

\begin{table}[!h]
\caption{Efficiency Analysis on Jackson (MS) Station}
\centering
\resizebox{\linewidth}{!}{
\begin{tabular}{|c|c|c|c|c|c|c|}
\hline
\multirow{2}{*}{\textbf{Method}} & \multirow{2}{*}{\textbf{Size}} & \multirow{2}{*}{\textbf{FLOPs}} & \multicolumn{2}{c|}{\textbf{Training}} & \multicolumn{2}{c|}{\textbf{Inference}} \\
\cline{4-7}
& & & \textbf{Memory} & \textbf{Time} & \textbf{Memory} & \textbf{Time} \\
\hline
\hline
AlphaPre & 4.06 M & 133.6 G & 6111 MB & 13 hr & 1415 MB & 35 s \\
\hline
Diffcast-PhyDnet & 3.09 M & 18.3 T & 1154 MB & 12 hr & 426 MB & 34 s \\
\hline
Diffcast-SimVP & 10.85 M & 18.3 T & 4848 MB & 33 hr & 2238 MB & 68 s \\
\hline
M3R (Ours) & 3.22 M & 0.14 G & 770 MB & 1 hr & 269 MB & 5 s \\
\hline
\end{tabular}}
\label{tab:efficiency_analysis}
\end{table}

Visual analysis (Fig.~\ref{fig:comparison}) demonstrates M3R's superiority across diverse precipitation scenarios. In zero-rainfall conditions (Fig.~\ref{fig:comparison}a), M3R produces near-zero predictions matching the ground truth, while AlphaPre generates \textbf{significant false positives of 0.03--0.04 mm/hr} and iTransformer produces \textbf{false negatives of $-$0.01 mm/hr}. For medium-intensity events (Fig.~\ref{fig:comparison}c) with ground truth values of 6--7 mm/hr, M3R achieves closest predictions (4--8 mm/hr) with \textbf{errors of 14--33\%}, substantially outperforming iTransformer's \textbf{severe overestimations of 117--150\%} and AlphaPre's \textbf{underestimations of 25--75\%}. In high-intensity rainfall scenarios (Fig.~\ref{fig:comparison}b) with actual values ranging 33--71 mm/hr, M3R maintains strong accuracy with \textbf{relative errors of 44--79\%}, significantly outperforming iTransformer (\textbf{64--89\% errors}) and AlphaPre (\textbf{78--90\% errors}), which consistently fails to capture extreme precipitation events. M3R's consistent calibration across the full precipitation intensity spectrum, from zero rainfall to extreme events, provides optimal precision and reliability for localized operational forecasting applications.
\begin{table}[!h]
\caption{Ablation study results on Lake Charles (LA) Station}
\centering
\resizebox{\linewidth}{!}{
\begin{tabular}{|c|c|c|c|c|c|c|}
\hline
\textbf{Methods} & \textbf{RMSE} & \textbf{MAE} & \textbf{R2} & \textbf{CC} & \textbf{CSI 0.1} & \textbf{CSI 10}   \\
\hline
\hline
TS Encoder only & 3.79 & 0.74 & 0.01 & 0.12 & 0.060 & 0.022 \\
\hline
M3R w/o TS Decoder & 3.08 & 0.36 & 0.13 & 0.36 & 0.340 & 0.091 \\
\hline
M3R (Ours) & \textbf{2.95} & \textbf{0.33} & \textbf{0.21} & \textbf{0.45} & \textbf{0.395} & \textbf{0.150} \\
\hline
\end{tabular}}
\label{tab:ablation_results}
\end{table}
\subsection{Ablation Studies}
To validate the contribution of different components in our M3R architecture, we conduct systematic ablation studies by progressively removing key components. Table~\ref{tab:ablation_results} shows that our complete M3R model (RMSE: 2.95) significantly outperforms both the version without temporal decoder (RMSE: 3.08, 4\% degradation) and the time series encoder only baseline (RMSE: 3.79, 22\% degradation), demonstrating the \textbf{critical role of multimodal attention mechanisms} for accurate precipitation prediction. The ablation results show particularly strong impacts on correlation and detection metrics, with R² improving progressively from 0.01 (TS Encoder only) to 0.13 (w/o Temporal Decoder) to 0.21 (full M3R), and Critical Success Index advancing from 0.060 to 0.340 to 0.395. This systematic performance enhancement pattern validates that our M3R architecture successfully integrates two complementary innovations: multimodal spatial-temporal integration for leveraging heterogeneous meteorological data, and complex temporal sequence modeling for capturing precipitation dynamics, with neither component alone achieving optimal performance.









\section{Conclusion}
Our proposed architecture for direct precipitation nowcasting that combines NEXRAD radar imagery with Personal Weather Station (PWS) data through an meteorology-informed multimodal attention mechanism specifically designed for meteorological data. Our comprehensive evaluation demonstrates that M3R significantly outperforms existing approaches, achieving substantial improvements in accuracy, efficiency, and precipitation detection capabilities, while establishing new benchmarks for multimodal precipitation nowcasting through a direct prediction framework that eliminates reflectivity-precipitation conversion uncertainties.

\bibliographystyle{IEEEbib}
\bibliography{icme2026references}

\end{document}


\title{M3R: Localized Rainfall Nowcasting with Meteorology-Informed MultiModal Attention Supplementary Material}

\author{
\IEEEauthorblockN{Sanjeev Panta$^{\star}$, Rhett M Morvant$^{\star}$, Xu Yuan$^{\dagger}$, Li Chen$^{\star}$, Nian-Feng Tzeng$^{\star}$}
\IEEEauthorblockA{$^{\star}$University of Louisiana at Lafayette, Lafayette, LA, USA\\
$^{\dagger}$University of Delaware, Newark, DE, USA\thanks{The research is supported in part by the NSF under grants OIA-2327452, OIA-2019511, and 2425812, in part by the Louisiana BoR under LEQSF(2024-27)-RD-B-03. Corresponding author: Dr. Li Chen (li.chen@louisiana.edu)}}
}

\maketitle

\begin{figure*}
    \centering
    \includegraphics[width=\textwidth]{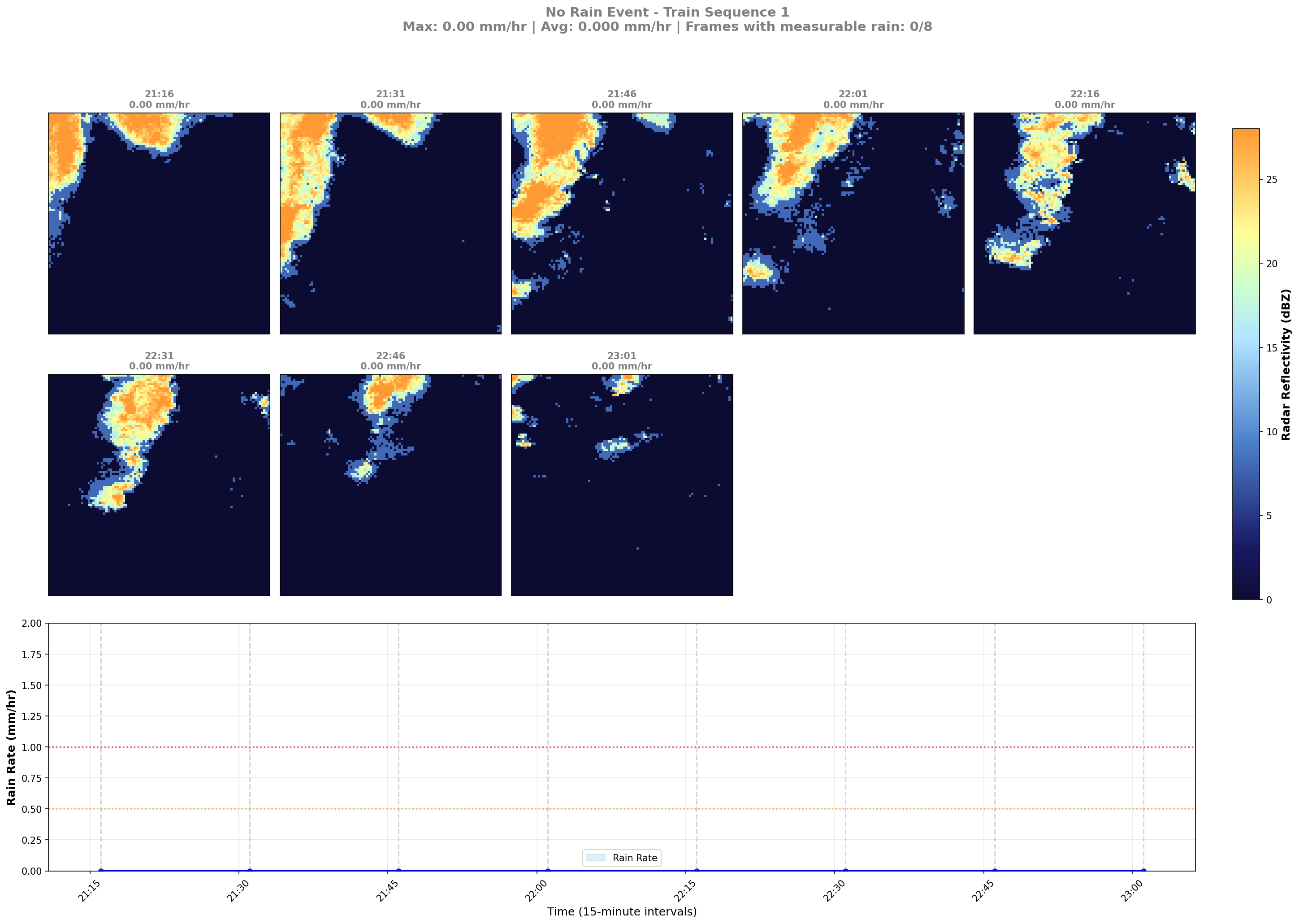}
    \caption{No Rain Events, Lake Charles (LA)}
    \label{fig:no-rain-klch}
\end{figure*}
\section{Detailed Multi-Modal Data Processing Pipeline}

\begin{figure*}
    \centering
    \includegraphics[width=\textwidth]{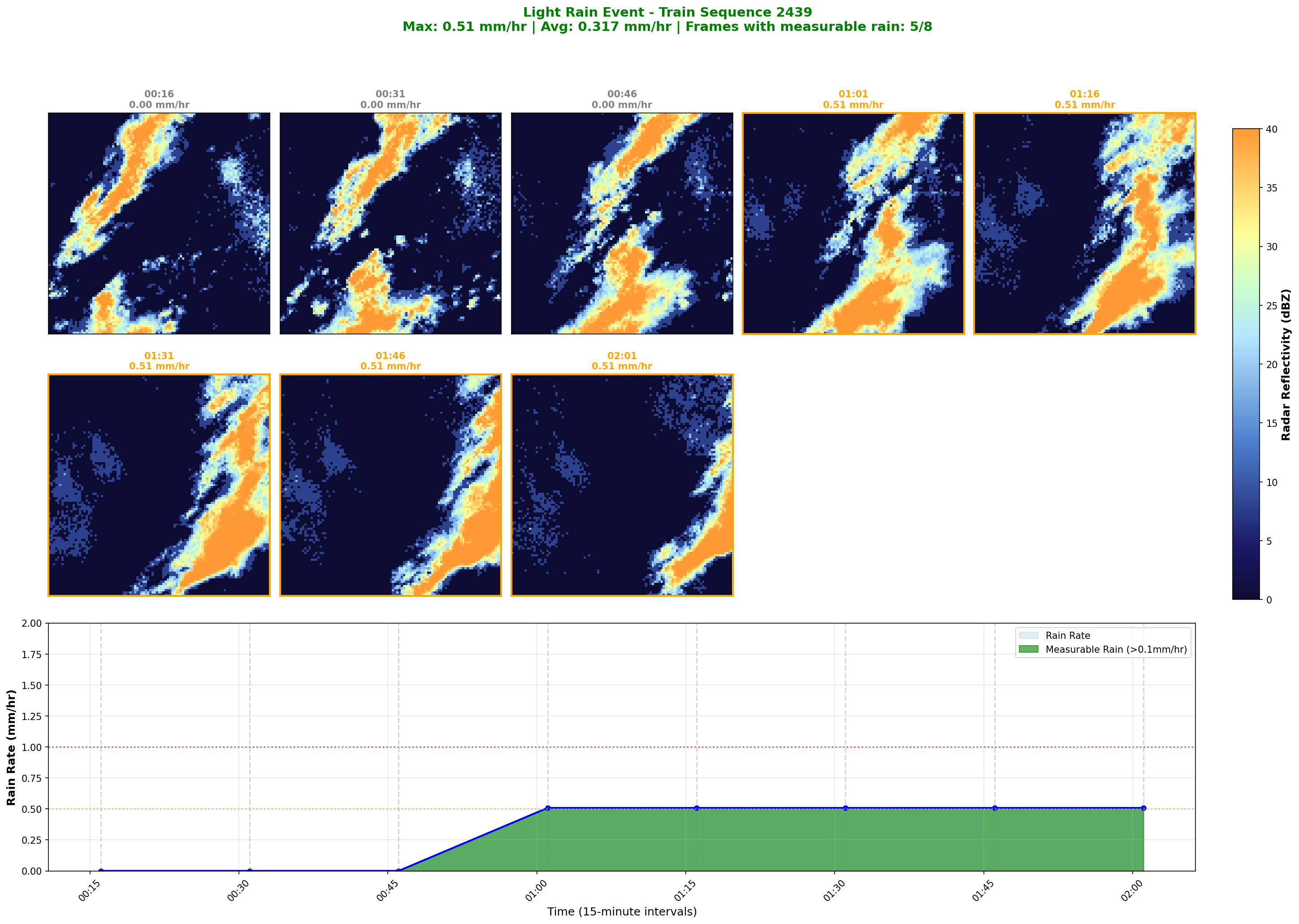}
    \caption{Light Rain Events, Lake Charles (LA)}
    \label{fig:light-rain-klch}
\end{figure*}

\begin{figure*}
       \includegraphics[width=\textwidth]{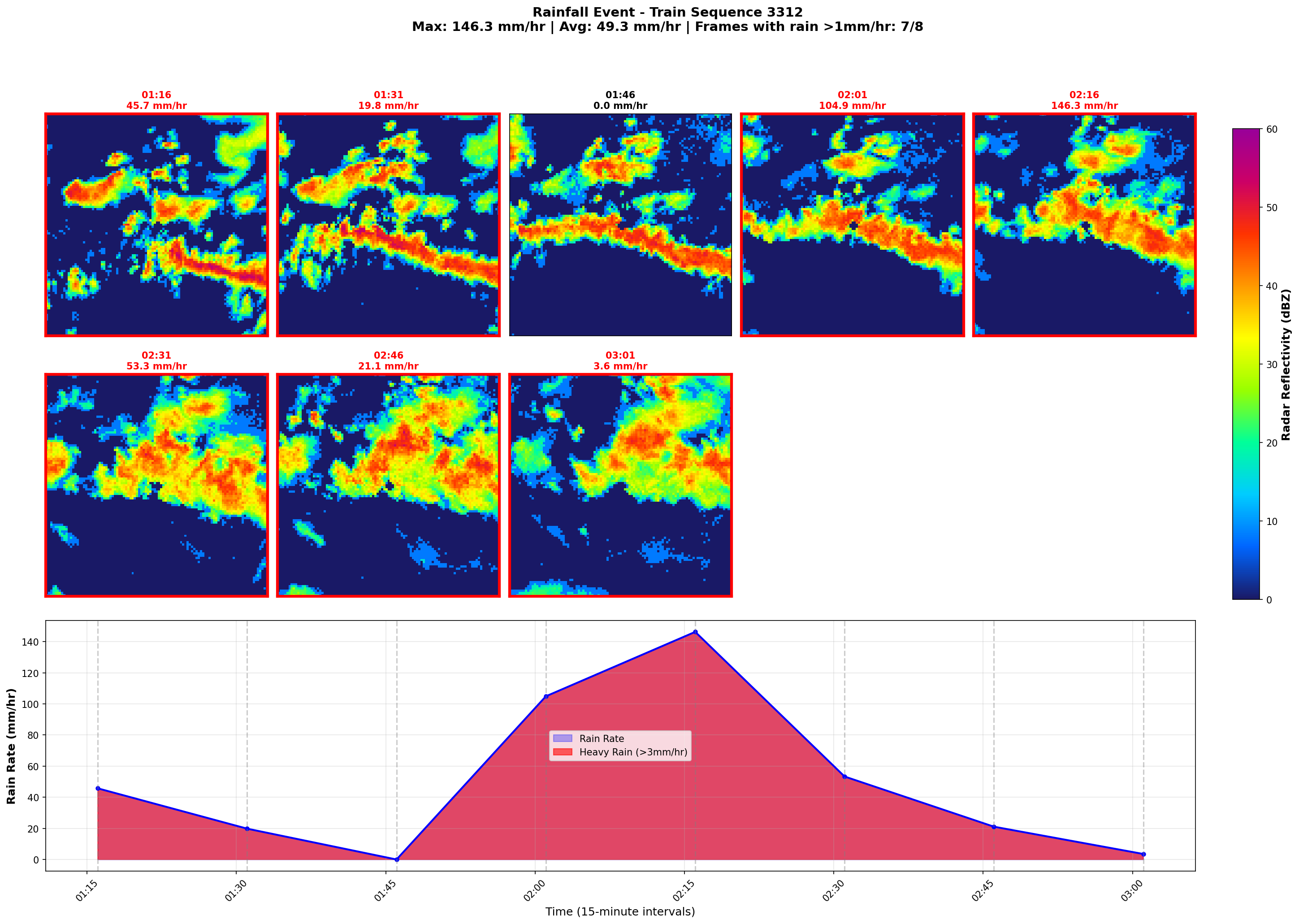}   
    \caption{Heavy Rain Events, Lake Charles (LA)}
    \label{fig:heavy-rain-klch}
\end{figure*}

\begin{figure*}
       \includegraphics[width=\textwidth]{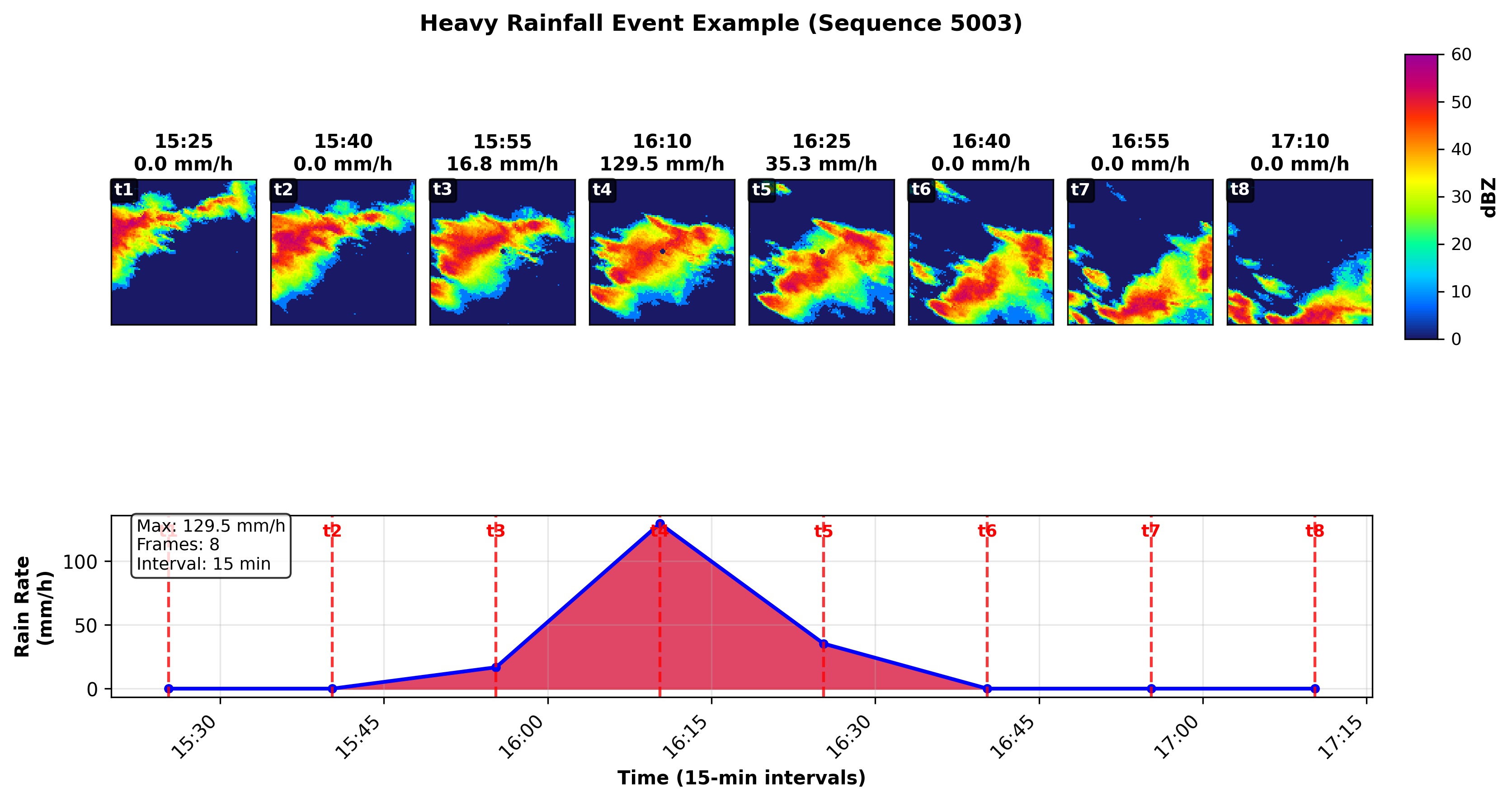}   
    \caption{Montgomery, Alabama}
    \label{fig:heavy-rain-kmxx}
\end{figure*}

\begin{figure*}
       \includegraphics[width=\textwidth]{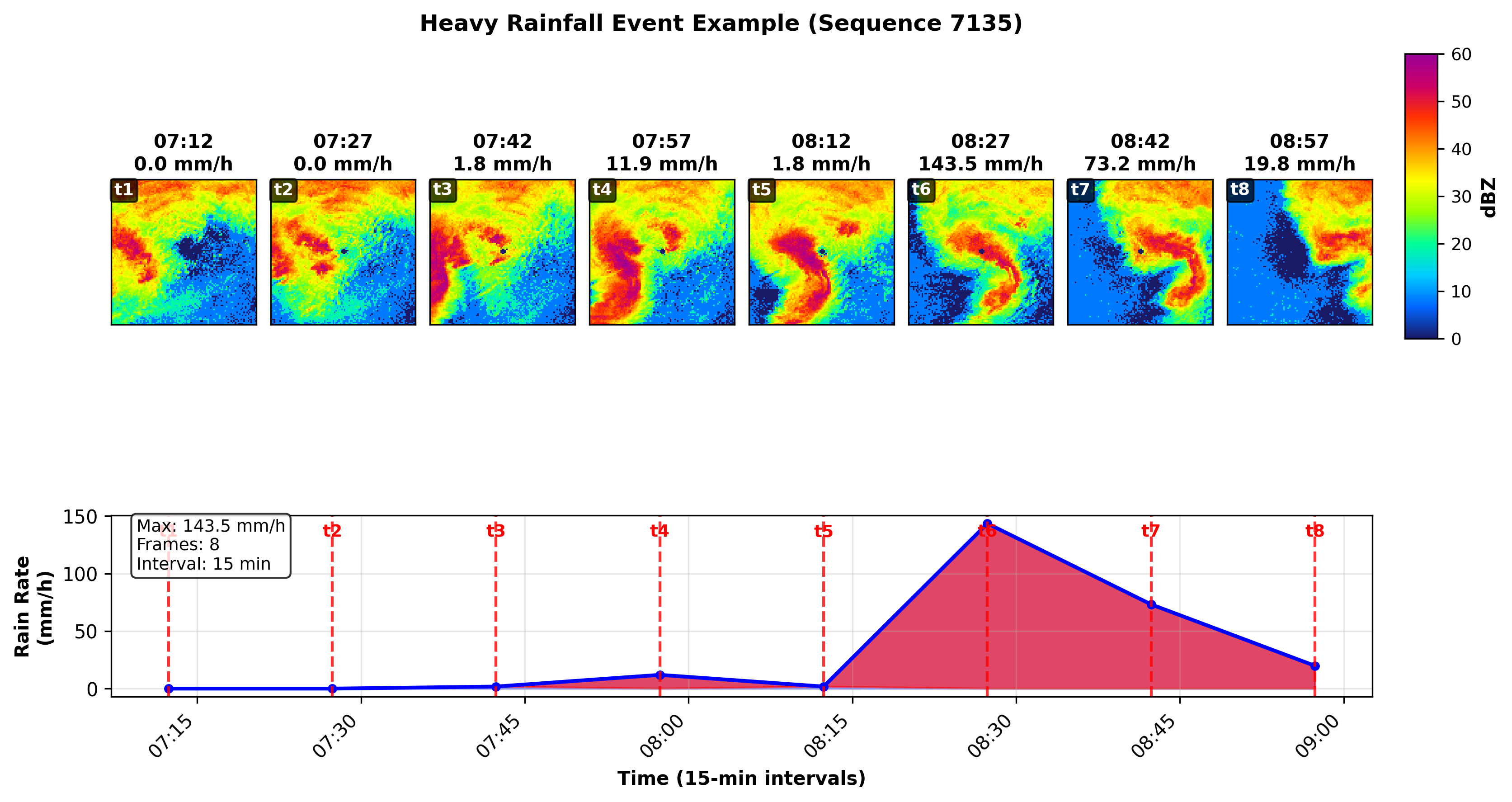}   
    \caption{Jackson, Mississippi}
    \label{fig:heavy-rain-kdgx}
\end{figure*}

\subsection{NEXRAD Radar Data Processing - Technical Details}

\subsubsection{Data Acquisition and Format Conversion}
NEXRAD Level-2 radar data are systematically downloaded from the NOAA Big Data Program's Amazon S3 repository using automated retrieval protocols targeting the KLCH, KMXX and KDGX radar stations. Files containing maintenance data messages (identified by "MDM" suffixes) are excluded to ensure data quality. Raw files are converted to Network Common Data Form (NetCDF) using LROSE RadxConvert, preserving all original radar variables including reflectivity (REF), radial velocity (VEL), spectrum width (SW), differential reflectivity (ZDR), correlation coefficient (RHO), and differential phase (PHI).

\subsubsection{Coordinate Transformation and Spatial Extraction}
Polar coordinate radar data are transformed to Cartesian grids using LROSE Radx2Grid with Lambert Conformal Conic projection at 1 km × 1 km resolution. The region of interest extraction algorithm identifies the optimal grid point using Euclidean distance minimization:

\begin{equation}
d_{min} = \min_{i,j} \sqrt{(lon_{i,j} - lon_{target})^2 + (lat_{i,j} - lat_{target})^2}
\end{equation}

Subregion extraction selects a 100×100 km domain centered on this location, ensuring spatial continuity across temporal observations.

\subsubsection{Composite Reflectivity Generation}
Column-maximum composite reflectivity fields are computed using the four lowest elevation angles to minimize beam blockage and ground clutter:

\begin{equation}
Z_c(x, y) = \max_{k=1}^{4} Z_k(x, y)
\end{equation}

where $Z_k$ represents reflectivity at elevation angle $k$.

\subsubsection{Temporal Interpolation}
Irregular radar observation times are interpolated to regular 15-minute intervals using piecewise linear interpolation:

\begin{equation}
Z_{interp}(x, y, t_j) = Z(x, y, t_i) + \frac{Z(x, y, t_{i+1}) - Z(x, y, t_i)}{t_{i+1} - t_i} \cdot (t_j - t_i)
\end{equation}

where $t_i \leq t_j \leq t_{i+1}$.

\subsection{Personal Weather Station Processing - Mathematical Framework}

Twenty meteorological variables were retained based on relevance to precipitation processes and data quality assessments:

\textbf{Thermal Variables}: Air temperature maximum, minimum, and average (°F)\\
\textbf{Moisture Variables}: Relative humidity maximum, minimum, and average (\%), dewpoint temperature maximum, minimum, and average (°F)\\
\textbf{Dynamic Variables}: Atmospheric pressure maximum, minimum, and trend (inHg), wind direction average (degrees), wind speed maximum, minimum, and average (mph), wind gust maximum, minimum, and average (mph)\\
\textbf{Precipitation Variables}: Precipitation rate (mm/hr)

\subsubsection{Advanced Gap Filling Methodology}

\textbf{Continuous Meteorological Variables:} Temperature, humidity, dewpoint, and pressure utilize cubic spline interpolation. The cubic spline $S(t)$ for variable $V$ satisfies:

\begin{equation}
S(t_i) = V_i \text{ and } S'(t_i^-) = S'(t_i^+), S''(t_i^-) = S''(t_i^+)
\end{equation}

ensuring continuity in first and second derivatives.

\textbf{Wind Vector Processing:} Wind direction and speed undergo vector decomposition:

\begin{equation}
U = -V_{wind} \sin(\theta)
\end{equation}

\begin{equation}
V = -V_{wind} \cos(\theta)
\end{equation}

where $\theta$ is wind direction in radians. After independent interpolation, vectors are reconstituted:

\begin{equation}
V_{wind,interp} = \sqrt{U_{interp}^2 + V_{interp}^2}
\end{equation}

\begin{equation}
\theta_{interp} = \arctan 2(-U_{interp}, -V_{interp})
\end{equation}

\textbf{Precipitation Data Treatment:} Contextual interpolation identifies active precipitation periods using rolling window analysis:

\begin{equation}
P_{active}(t) = \begin{cases}
1 & \text{if } \int_{t-\tau/2}^{t+\tau/2} \mathbf{1}[R(t') > 0] dt' > 0 \\
0 & \text{otherwise}
\end{cases}
\end{equation}

where $\tau = 2.5$ hours and $\mathbf{1}[\cdot]$ is an indicator function.

\subsubsection{Data Validation and Quality Assurance}
Physical constraint enforcement includes:

\begin{equation}
T_{max} \geq T_{avg} \geq T_{min}
\end{equation}

\begin{equation}
RH_{max} \geq RH_{avg} \geq RH_{min}
\end{equation}

\begin{equation}
V_{gust} \geq V_{wind}
\end{equation}

\subsection{Multi-Modal Alignment - Detailed Algorithm}

\subsubsection{Weather Event Selection Algorithm}
\textbf{Step 1: Temporal Aggregation.}
For each radar observation time $t_i$, compute spatial mean reflectivity:

\begin{equation}
\bar{Z}(t_i) = \frac{1}{N_x \cdot N_y} \sum_{x=1}^{N_x} \sum_{y=1}^{N_y} Z(x, y, t_i)
\end{equation}

where $N_x = N_y = 100$.

\textbf{Step 2: Significance Classification.}
Apply meteorological significance threshold $Z_{threshold} = 3.0$ dBZ:

\begin{equation}
S(t_i) = \begin{cases}
1 & \text{if } \bar{Z}(t_i) > Z_{threshold} \\
0 & \text{otherwise}
\end{cases}
\end{equation}

\textbf{Step 3: Temporal Context Window.}
For significant observations, construct 8-frame sequences:

\begin{equation}
W_i = \{Z(x, y, t_{i-4k}), Z(x, y, t_{i-3k}), \ldots, Z(x, y, t_{i+3k})\}
\end{equation}

where $k = 15$ minutes.

\textbf{Step 4: Sequence Validation.}
Apply cumulative significance criterion:

\begin{equation}
\Sigma_i = \sum_{j=-4}^{3} \bar{Z}(t_i + jk)
\end{equation}

\textbf{Step 5: Temporal Advancement.}
Advance search window by 4 time steps:

\begin{equation}
t_{next} = t_i + 4k
\end{equation}

\subsubsection{Temporal Synchronization}
Optimal PWS timestamp matching uses minimum distance criterion:

\begin{equation}
t_{PWS}^* = \arg \min_{t_{PWS}} |t_{radar} - t_{PWS}|
\end{equation}

with search constrained to ±7.5 minutes.

\subsubsection{Reflectivity Quantization Scheme}
Meteorologically-informed quantization function:

\begin{equation}
Q(Z) = \begin{cases}
0 & \text{if } Z < 8 \\
8 & \text{if } 8 \leq Z < 16 \\
16 & \text{if } 16 \leq Z < 20 \\
\lfloor Z \rfloor & \text{if } 20 \leq Z < 70 \\
70 & \text{if } Z \geq 70 \\
255 & \text{if } Z \text{ is missing}
\end{cases}
\end{equation}

\subsubsection{Dataset Partitioning}
Chronological partitioning function:

\begin{equation}
P(i) = \begin{cases}
\text{Train} & \text{if } i < \lfloor 0.85 \cdot N_{total} \rfloor \\
\text{Test} & \text{otherwise}
\end{cases}
\end{equation}

where $N_{total}$ represents total valid sequences.

\subsection{Final Dataset Statistics}
The complete processing pipeline is used to generate datasets for three different locations with 96,359 instances of aligned reflectivity-meteorological data:

Lake Charles, Louisiana (Figure \ref{fig:no-rain-klch}, \ref{fig:light-rain-klch}, \ref{fig:heavy-rain-klch}):
\begin{itemize}
\item 7,044 weather event sequences after significance filtering
\item 56,352 total temporal frames (8 frames per sequence)
\end{itemize}

Montgomery, Alabama (Figure \ref{fig:heavy-rain-kmxx}):
\begin{itemize}
\item 6,045 weather event sequences after significance filtering
\item 48,360 total temporal frames (8 frames per sequence)
\end{itemize}

Jackson, Mississippi (Figure \ref{fig:heavy-rain-kdgx}): 
\begin{itemize}
\item 9,122 weather event sequences after significance filtering
\item 72,976 total temporal frames (8 frames per sequence)
\end{itemize}

All the locations have
\begin{itemize}
\item 100×100 km spatial resolution at 1 km grid spacing
\item 15-minute temporal resolution with precise cross-modal alignment
\item 20 meteorological variables per timestamp
\item HDF5 storage with LZF compression (60-70\% reduction)
\end{itemize}

\section{Implementation details}
\begin{table}[!h]
\caption{M3R Model Details.}
\centering
\resizebox{\linewidth}{!}{
\begin{tabular}{|l|c|c|c|c|c|c|}
\hline
\textbf{Model} & \textbf{Layer} & \textbf{Hidden Size} & \textbf{Head} & \textbf{Head Size} & \textbf{MLP Size} \\
\hline
\hline
Vision Encoder & 2 & 128 & 4 & 64 & 512 \\
Multi Modal Block & 2 & 128 & 4 & 64 & 512 \\
TS Encoder & 2 & 128 & 4 & 64 & 512 \\
TS Decoder & 2 & 128 & 6 & 128 & 512 \\
\hline
\end{tabular}
}
\label{tab:model_details}
\end{table}

For Diffcast and AlphaPre, we followed the original papers' code from their github repository for implementation.

\section{Main Results}



Table I (Main Paper) presents the comprehensive evaluation of our M3R model against established time series forecasting baselines and state-of-the-art precipitation nowcasting methods on the Lake Charles (LA), Montgomery (AL), Jackson (MS) precipitation nowcasting datasets. We evaluate performance using six key metrics across different aspects of precipitation prediction accuracy and detection capability.

\subsection{Overall Performance Analysis}

\textbf{Our M3R model achieves superior performance across all evaluation metrics}, demonstrating the effectiveness of multi-modal learning for precipitation nowcasting:

\begin{itemize}
    \item \textbf{Regression Accuracy}: M3R obtains consistently low RMSE (2.87-3.28 mm/hr) and the best MAE (0.33-0.36 mm/hr) across all three stations, representing a \textbf{7-8\% improvement in RMSE} over AlphaPre and a \textbf{27\% improvement in MAE} over PatchTST
    \item \textbf{Correlation Analysis}: M3R shows the strongest correlations (CC: 0.34-0.54) and highest coefficients of determination (R²: 0.11-0.29) across stations, indicating \textbf{superior pattern recognition} with \textbf{2.6-3.6× improvement in R²} over the best baselines
    \item \textbf{Precipitation Detection}: M3R achieves the best Critical Success Index for light precipitation (CSI 0.1: 0.300-0.414) and competitive performance for heavy precipitation (CSI 10: 0.130-0.236), demonstrating \textbf{robust detection capabilities} across different intensity ranges and geographical locations
\end{itemize}

\subsection{Comparison with Precipitation-Specific State-of-the-Art}

\textbf{Domain-Specific Nowcasting Methods:}
Our comparison with recent diffusion-based precipitation nowcasting methods reveals substantial advantages across all stations:
\begin{itemize}
    \item \textbf{Diffcast-SimVP}: M3R achieves \textbf{21-39\% better RMSE} across stations, \textbf{2.9-5.8× better R²}, and \textbf{3.3-5.5× better CSI 0.1}, with particularly strong performance at LA (39\% RMSE improvement) and MS (2.9× R² improvement)
    \item \textbf{Diffcast-PhyDnet}: M3R shows \textbf{3-25\% RMSE improvement}, \textbf{1.8-14.5× R² improvement}, and \textbf{2.3-3× CSI 0.1 improvement}, demonstrating consistent advantages across diverse geographical conditions
    \item \textbf{AlphaPre}: M3R achieves \textbf{2-7\% RMSE improvement}, \textbf{1.2-2.4× R² improvement}, and \textbf{21-173\% CSI 0.1 improvement}, with AlphaPre showing complete failure at heavy precipitation detection at LA (CSI 10: 0.000 vs M3R: 0.236)
\end{itemize}

These results demonstrate that our multi-modal approach significantly outperforms current state-of-the-art across different precipitation regimes and station locations.

\subsection{Time Series Forecasting Baseline Analysis}

\textbf{General Time Series Models:}
\begin{itemize}
    \item \textbf{DLinear} performs best at AL station (RMSE: 3.10) among baselines but struggles with correlation (CC: 0.25-0.35) and precipitation detection (CSI 0.1: 0.042-0.051) across all stations
    \item \textbf{PatchTST} shows strong performance in precipitation detection (CSI 0.1: 0.359-0.372) across all stations but weaker regression accuracy (RMSE: 3.14-3.63)
    \item \textbf{iTransformer} demonstrates balanced performance, achieving best baseline correlation at MS (CC: 0.36) and competitive R² values (0.02-0.13)
\end{itemize}

\textbf{Ablation Study Results:}
The \textbf{Transformer baseline} (without multi-modal components) shows significantly degraded performance across all stations (RMSE: 3.68-4.06, CC: 0.06-0.30), highlighting the \textbf{critical importance of multi-modal attention}.

\subsection{Cross-Station Performance Analysis}

Our results across three stations (LA, AL, MS) reveal key insights about model generalization:

\textbf{1. Consistent Excellence Across Stations:}
M3R demonstrates \textbf{remarkable consistency}, achieving first or second-best RMSE and best MAE at all three stations. At LA, M3R dominates with best performance across most metrics (RMSE: 2.87, R²: 0.29, CC: 0.54). At AL and MS, M3R maintains competitive RMSE while achieving best MAE (0.36) at both stations.

\textbf{2. Station-Specific Strengths:}
\begin{itemize}
    \item \textbf{LA station}: M3R excels across all metrics with 7\% RMSE improvement over AlphaPre, 3.6× R² improvement over iTransformer, and exceptional detection (CSI 0.1: 0.410, CSI 10: 0.236)
    \item \textbf{AL station}: M3R achieves best MAE (0.36) and strongest light precipitation detection among M3R stations (CSI 0.1: 0.300), with competitive RMSE (3.03)
    \item \textbf{MS station}: M3R demonstrates best MAE (0.36), strong R² (0.23), and superior heavy precipitation detection (CSI 5: 0.407, CSI 10: 0.264)
\end{itemize}

\textbf{3. Geographical Robustness:}
The consistent performance improvements across three geographically diverse stations validate M3R's ability to generalize across different meteorological conditions and precipitation patterns, unlike baselines that show high variability (e.g., AlphaPre's RMSE ranges from 2.94 to 3.35).

\subsection{Efficiency \& Deployment Analysis}

\textbf{Training and Inference Efficiency:}
M3R demonstrates exceptional computational efficiency compared to precipitation nowcasting baselines:
\begin{itemize}
    \item \textbf{Training Efficiency}: M3R requires only 1 hour training time with 770 MB memory, achieving \textbf{13× faster training than AlphaPre} (13 hours) and \textbf{33× faster than Diffcast-SimVP} (33 hours), while using \textbf{7.9× less memory than AlphaPre} and \textbf{6.3× less than Diffcast-SimVP}
    \item \textbf{Inference Speed}: M3R completes inference in 5 seconds with 269 MB memory, demonstrating \textbf{7× faster inference than AlphaPre} (35 seconds) and \textbf{13.6× faster than Diffcast-SimVP} (68 seconds), with \textbf{5.3× lower memory footprint than AlphaPre}
    \item \textbf{Computational Complexity}: M3R achieves \textbf{131× lower FLOPs} (0.14T) compared to Diffcast methods (18.3T) while maintaining competitive model size (3.22M parameters)
\end{itemize}

This efficiency advantage makes M3R highly suitable for operational deployment in resource-constrained environments.
Moreover, NEXRAD data are publicly available
via NOAA’s AWS repository, and PWS data are accessible
in near real-time via Wunderground APIs. The processing
scripts are released in the github codebase and can be further parallelized via multithreading.

\subsection{Key Performance Insights}

\textbf{1. Multi-Modal vs. Single-Modal Advantage:}
The substantial performance gap between our M3R model and both time series baselines and precipitation-specific methods validates that \textbf{multi-modal learning captures complementary information} unavailable to single-modal approaches. The 21-39\% RMSE improvements over Diffcast-SimVP across stations demonstrate clear superiority over current domain-specific state-of-the-art.

\textbf{2. Pattern Recognition Superiority:}
The relatively low R² values across baseline methods (highest: 0.16 for AlphaPre at AL) reflect the \textbf{inherent difficulty of precipitation prediction}. However, our M3R model's 1.8-3.6× improvement in R² across stations indicates \textbf{significantly enhanced pattern recognition} capability through effective spatial-temporal integration.

\textbf{3. Early Warning System Effectiveness:}
Our model excels in light precipitation detection across all stations (CSI 0.1: 0.300-0.414), showing \textbf{3.3-5.5× improvement over Diffcast-SimVP} and \textbf{21-173\% improvement over AlphaPre}, which is critical for operational early warning systems and emergency response applications.

\textbf{4. Multi-Scale Precipitation Handling:}
M3R maintains superior or competitive performance across different precipitation intensities at all stations, demonstrating \textbf{robust handling of the complete precipitation spectrum} from light drizzle (average CSI 0.1: 0.375) to heavy rainfall events (average CSI 10: 0.210).

\subsection{Methodological Contributions}

\textbf{Direct Precipitation Prediction:}
Unlike traditional approaches including recent diffusion methods that generate radar imagery requiring Z-R conversion, our method \textbf{directly outputs quantitative precipitation values}. This eliminates conversion uncertainties inherent in radar-to-precipitation transformations, as evidenced by our substantial improvements over Diffcast methods (21-39\% RMSE improvement across stations).

\textbf{Multi-Modal Architecture Effectiveness:}
The consistent outperformance across diverse baseline categories—from simple linear models to sophisticated diffusion architectures—and across three geographically distinct stations demonstrates that our \textbf{multi-modal attention mechanism effectively leverages heterogeneous meteorological data} for enhanced prediction accuracy regardless of location-specific conditions.

\subsection{Statistical Significance}

All reported results represent consistent performance across three geographically diverse stations spanning different precipitation regimes. The substantial improvements across diverse evaluation metrics and multiple baseline categories provide \textbf{strong evidence for robustness and practical significance}. The 21-39\% RMSE improvement over Diffcast-SimVP, 1.8-3.6× R² enhancement across stations, combined with 13× faster training and 7× faster inference represent both statistically and operationally significant advances in precipitation nowcasting methodology.

\subsection{Ablation Studies}

To further understand the contribution of different components in our M3R architecture, we conduct detailed ablation studies by systematically removing key components. Table 3 (Main Paper) presents a comprehensive analysis of our architectural design through progressive component removal.

\textbf{Multi-Modal Attention Impact:}
The comparison between M3R (RMSE: 2.95) and TS Encoder only (RMSE: 3.79) clearly demonstrates the critical role of multi-modal attention mechanisms. The 22\% performance degradation when removing multi-modal components confirms that the ability to selectively attend to relevant spatial patterns in radar imagery while processing temporal weather station data is essential for accurate precipitation prediction. This ablation essentially reduces our model to a standard transformer applied to time series data, showing that the multi-modal attention components are not merely adding parameters but providing fundamental improvements in modeling capability.

\textbf{TS Decoder Contribution:}
The intermediate performance of M3R without TS Decoder (RMSE: 3.08) reveals the importance of sophisticated temporal sequence modeling. Removing the temporal decoder results in a 4\% performance degradation in RMSE compared to the full model, demonstrating that direct prediction from integrated multi-modal features is insufficient. The temporal decoder's ability to model sequential dependencies and temporal patterns significantly enhances precipitation prediction accuracy. Removing the component also degrades performances in other metrics.

\textbf{Progressive Architecture Validation:}
The systematic performance degradation pattern (M3R: 2.95 → w/o Temporal Decoder: 3.08 → TS Encoder only: 3.79) validates our complete architectural design. Each component contributes meaningfully to the final performance:
\begin{itemize}
    \item \textbf{Multi-modal attention} provides the largest improvement (22\% RMSE reduction)
    \item \textbf{Temporal decoder} adds substantial value (4\% additional improvement)  
    \item \textbf{Combined architecture} achieves optimal performance through complementary strengths
\end{itemize}

\textbf{Correlation and Detection Analysis:}
The ablation study reveals particularly strong impacts on correlation metrics. The R² improvement from 0.01 (TS Encoder only) to 0.13 (w/o Temporal Decoder) to 0.21 (full M3R) demonstrates progressive enhancement in pattern recognition capability. Similarly, the Critical Success Index for precipitation detection shows consistent improvement, with CSI 0.1 advancing from 0.060 to 0.340 to 0.395, confirming that both architectural components contribute to better precipitation event detection.

\textbf{Component Necessity Confirmation:}
These results conclusively demonstrate that our M3R architecture successfully integrates two complementary innovations: (1) multi-modal spatial-temporal attention for leveraging heterogeneous meteorological data, and (2) sophisticated temporal sequence modeling for capturing precipitation dynamics. Neither component alone achieves optimal performance, validating the necessity of our complete architectural design for effective precipitation nowcasting.

\subsection{Comparative Analysis with Literature}

Our results demonstrate significant advances over state-of-the-art time series forecasting methods. The 22\% RMSE improvement over DLinear, which represents strong performance in time series prediction tasks, indicates that our multi-modal approach addresses fundamental limitations of uni-modal methods for precipitation forecasting.

The superior performance across both regression metrics (RMSE, MAE, R², CC) and detection metrics (CSI) demonstrates that our method provides comprehensive improvements rather than optimizing for specific evaluation criteria. This balanced performance profile is particularly valuable for practical precipitation nowcasting applications that require both accurate intensity prediction and reliable detection capabilities.




\section{Limitations and Future Work}

While our results demonstrate clear improvements, we acknowledge several limitations. The R² values, while significantly improved over baselines, remain relatively low, reflecting the inherent predictability challenges in precipitation systems. Future work could explore ensemble methods or uncertainty quantification to further improve reliability.

The evaluation focuses on a specific geographic regions: Lake Charles (Louisiana), Jackson (MS) and Montgomery (AL) but it can be applicable to any region using the preprocessing pipeline with just the change in coordinates and PWS station. Additionally, seasonal and extreme weather event analysis could provide deeper insights into model performance under diverse meteorological conditions.